\documentclass[conference]{IEEEtran}

\usepackage{cite}
\usepackage{color}

\newif\ifdraft
\draftfalse
\drafttrue
\ifdraft
 \newcommand{\jhanote}[1]{ {\textcolor{red} { ***SJ: #1 }}}
 \newcommand{\gcfnote}[1]{ {\textcolor{blue} { ***WE: #1 }}}
 \else
 \newcommand{\jhanote}[1]{}
 \newcommand{\gcfnote}[1]{}
\fi

\usepackage{url}
\usepackage{paralist}

\begin{document}


\title{Understanding ML driven HPC: Applications and Infrastructure}

\author{Geoffrey Fox$^{1}$, Shantenu Jha$^{2}$$^{,3}$\\

\small{\emph{$^{1}$ Indiana University, Bloomington, IN}}\\
\small{\emph{$^{2}$ Rutgers, State University of New Jersey, Piscataway, NJ
08854, USA}}\\
\small{\emph{$^{3}$ Brookhaven National Laboratory, Upton, New York, 11973}}\\

}


\maketitle
\date{}
\begin{abstract}

We recently outlined the vision of  "Learning Everywhere" which captures the
possibility and impact of how learning methods and traditional HPC methods can
be coupled together. A primary driver of such coupling is the promise that
Machine Learning (ML) will give major performance improvements for traditional
HPC simulations. Motivated by this potential, the ML around HPC class of
integration is of particular significance. In a related follow-up paper, we
provided an initial taxonomy for integrating learning around HPC methods. In
this paper, which is part of the Learning Everywhere series, we discuss
``how'' learning methods and HPC simulations are being integrated to enhance
effective performance of computations. This paper identifies several modes ---
substitution, assimilation, and control, in which learning methods integrate
with HPC simulations and provide representative applications in each mode.
This paper discusses some open research questions and we hope will motivate
and clear the ground for MLaroundHPC benchmarks.

\end{abstract}

\section{Introduction and Motivation}

The convergence of HPC and learning methodologies provides a promising
approach to major performance improvements. Traditional HPC simulations are
reaching the limits of original progress. The end of Dennard scaling of
transistor power usage, and the end of Moore’s Law as originally formulated
has yielded fundamentally different processor architectures. The architectures
continue to evolve, resulting in highly costly, if not damaging churn in
scientific codes that need to be finely tuned to extract the last iota of
parallelism and performance. This approach to high-performance scientific
computing is simply unsustainable.

In domain sciences such as biomolecular sciences, advances in statistical
algorithms and runtime systems have enabled extreme scale ensemble based
applications~\cite{KASSON201887} to overcome limitations of traditional
monolithic simulations. However, in spite of several orders of magnitude
improvement in efficiency from these adaptive ensemble algorithms, the
complexity of phase space and dynamics for modest physical systems, require
additional orders of magnitude improvements and performance gains. Integrating
traditional HPC approaches with machine learning methods holds significant
promise towards overcoming these barriers.


It has always been necessary to improve the effectiveness of simulations;
however, its necessity and significance increases drastically at large-scales.
First, there is a need to enhance, if not preserve computational efficiency at
scale. Applying high-performance computing capabilities at (exa-)scale, leads
to the possibility of greater scientific inefficiency in computational
campaigns. For example, greater computational capacity might generate
relatively greater correlations and lower sampling, and thus less independent
data and exploration. Algorithms, methods and campaign strategies that worked
at lower scales are not necessarily suitable at greater scales. Second,
traditional computational campaigns have not exploited the intermediate data
from high-performance computing to their fullest: computational campaigns have
been conducted in a static, if not ad hoc fashion based upon initial
assumptions and states. The implications of static computational campaigns
will be exacerbated at scale, and thus novel algorithms, methods and campaign
strategies are needed that employ sophisticated learning to utilize and adapt
to intermediate data products.

In many application domains, the integration of ML into computations is a
promising way to obtain large performance gains, and presents an opportunity
to jump a generation of simulation enhancements. For example, one can view the
use of learned surrogates as a performance boost that can lead to huge
speedups, as calculation of a prediction from a trained network can be many
orders of magnitude faster than full execution of the
simulation~\cite{Fox2019Learning,Lamim_Ribeiro2019-qm}. In addition to the use
of learning for advanced sampling as illustrated above, simple examples are
the use of a surrogate to represent a chemistry potential, or a larger grain
size to solve the diffusion equation underlying cellular and tissue level
simulations.



This paper explores opportunities at the interface between high-performance
simulations and machine learning. Specifically, it investigates how ML driven
HPC simulations --- which based upon the taxonomy introduced in
Ref.~\cite{Fox2019Learning} is referred to as the ``ML around HPC'' --- can
pervasively enhance high-performance computational science. It attempts to
answer questions such as: How and where can ML effectively enhance HPC
simulations? When should ML methods substitute traditional simulations? Which
ML methods are promising? What are the general motifs or patterns of
interaction between ML and HPC?

In order to provide a quantitative metric by which to measure and answer some
of these questions, it is necessary to distinguish between traditional
performance measured by operations per second or benchmark scores, from the
effective performance that one gets by combining learning with simulation
which gives increased scientific performance --- as determined by a suitable
metric and measure, without changing the traditional system characteristics.

In general, there are three types of performance that require distinction: the
first, traditional system or application performance, which is measured by
typical scaling, utilization or operations by second and benchmark scores. The
second is the improvement in the computational investigation of the scientific
process, as measured by time-to-solution (for a given resource amount) or
another scientific metric. The third measure of performance, is the increase
in either the learning phase due to being trained with (physically
meaningfully) simulations, or the improvement in the simulation due to being
interaction with learning phase.

In cases where there is a (tight) coupling between the learning and simulation
components, the second measure of performance is of paramount importance. It
motivates the notion of crossover point defined as the point in configuration
space at which the learning method is either more performant --- efficient
(e.g., same quality of results produced with less computing), or better (e.g.,
produces “better” results than possible via first principles / simulations),
or faster than simulations (e.g., speeding up or classic effective performance
a la reduced order modeling). Crossover points are dependent upon several
factors including the complexity of underlying model and problem, availability
of surrogate, the sensitivity to the ratio of cost / value of data (e.g, is it
better to have lots of cheap and low quality data, or small amount of high
quality data).


This paper is a follow on from Learning Everywhere~\cite{Fox2019Learning} and
an accompaniment to the article on ``Taxonomy of MLaroundHPC'' as part of the
IEEE eScience 2019. In Section II, we summarize the high-level organization of
MLforHPC, followed by an taxonomy of different MLaroundHPC applications. In
Section III, we focus on MLaroundHPC --- investigating the different modes,
mechanisms and functional motivations of integration of ML around HPC. We also
discuss some canonical examples of the different modes of integration. We will
use insights gained from an investigation of these issues to discuss open
issues and research challenges in cyberinfrastructure --- algorithms \&
methods, software and hardware, that must be addressed in the near and
intermediate term.

We thank the organizers of the IEEE eScience 2019 for the
opportunity to contribute to the Vision Track. We believe eScience conference
series has an important and distinguished track record of bringing the data
sciences -- methods and infrastructure, closer to traditional simulation based
science. We hope this article will help the eScience Steering Committee to
keep the conference series aligned with the thinking, needs and future
directions of the community to push the boundaries of computational and data
driven discovery.


\section{Learning Everywhere}

We have identified \cite{Fox2019Learning} several important distinctly
different links between machine learning (ML) and HPC. We term the full area
\textbf{MLandHPC} and define two broad categories \cite{Jeff_Dean2017}:
\textbf{HPCforML} and \textbf{MLforHPC}. HPCforML uses HPC to execute and
enhance ML performance, or using HPC simulations to train ML algorithms
(theory guided machine learning), which are then used to understand
experimental data or simulations. On the other hand \textbf{MLforHPC} uses ML
to enhance HPC applications and systems, where big data comes from the
computation and/or experimental sources that are coupled to ML and/or HPC
elements. MLforHPC can be further subdivided as  {\it MLaroundHPC}, {\it
MLControl}, {\it MLAutotuningHPC}, and {\it MLafterHPC} described in detail
below.



The MLforHPC category covers all aspects of machine learning interacting with
computation typically implemented as HPC. The sub-categories are useful but
incomplete, and definitely not always precise. There is a need to improve the
conceptual understanding of the different facets and dimensions of MLforHPC.
We delineate the initial types of MLforHPC we have identified:

\subsection{MLaroundHPC}  Using ML to learn from simulations and
produce learned surrogates for the simulations. This increases effective
performance for strong scaling where we keep the problem fixed but run it
faster or run the simulation for a longer time interval such as that relevant
for biological systems. It includes \textit{SimulationTrainedML} where the
simulations are performed to directly train an AI system rather than the AI
system being added to learn a simulation. Some common ways in which
MLaroundHPC is used, include:

\begin{asparaenum}

\item \textbf{MLaroundHPC: Learning Outputs from Inputs} Simulations performed to
directly train an AI system, rather than AI system being added to learn a
simulation~\cite{Noe2018-jc,Endo2018-ue}.

\item \textbf{MLaroundHPC: Learning Simulation Behavior} ML learns behavior
replacing detailed computations by ML
surrogates~\cite{Gentzsch2018,Han2018-il}

\item \textbf{MLaroundHPC: Faster and Accurate PDE Solutions} Efficient
numerical solution of PDEs is one of the most costly computations in many
simulations, and solving high dimensional PDEs such as the diffusion equation
has been notoriously difficult. Recent ML accelerated
algorithms~\cite{Han2018-il} for solving high-dimensional nonlinear PDEs are
effective for a wide variety of problems, in terms of both accuracy and speed.
These algorithms~\cite{Sirignano2018-li} approximate the solution
high-dimensional PDEs such as the diffusion equation using a ``Deep Galerkin
Method (DGM)'', and train their  network on batches of randomly sampled time
and space points. 
These new AI accelerated
approaches~\cite{Tripathy2018-fx,Raissi2017-ep} open up a host
of possibilities in materials, physics and cosmology, and scientific computing
more generally.


\item \textbf{MLaroundHPC: New Approach to Multi-Scale Modeling}
\emph{Effective potential} is an analytic, quasi-emperical or
quasi-phenomological potential that combines multiple, perhaps opposing,
effects into a single potential. For example,  we have a model specified at a
microscopic scale and we define a coarse graining to a different scale with
macroscopic entities defined to interact with effective dynamics specified in
some fashion such as an effective potential or effective interaction graph.
Machine learning is \emph{ideally} suited for defining effective potentials
and order parameter dynamics, and shows significant promise to deliver orders
of magnitude performance increases over traditional coarse-graining and order
parameter approaches. See well established methods
\cite{behler_generalized_2007,behler_first_2017,butler_machine_2018,s.smith_ani-1:_2017,s_smith_outsmarting_2018,smith_less_2018}

\end{asparaenum}


\subsection{MLControl} Two representative scenarios are:

\begin{asparaenum}
\item \textbf{Experiment Control}
Using simulations (possibly with HPC) in control of experiments and in objective driven computational campaigns \cite{Alexander2018}.  Here the simulation surrogates are very valuable to allow real-time predictions. 
Examples about: Material Science \cite{Yager2018-zw,Ren2018-gv,Ward2018-lo}, Fusion\cite{Bill_Tang2018-rg}, Nano \cite{balachandran2016adaptive}

\item \textbf{Experiment Design}
A big challenge is the uncertainty in the precise model structures and
parameters. Model-based design of experiments (MBDOE) assists in the planning
of highly effective and efficient experiments -- it capitalizes on the
uncertainty in the models to investigate how to perturb the real system to
maximize the information obtained from experiments. MBDOE with new 
ML assistance~\cite{snoek2012practical}
identifies the optimal conditions for stimuli and measurements that yield the
most information about the system given practical limitations on realistic
experiments.

\end{asparaenum}

\subsection{MLAutoTuning} Captures the scenario where ML is used to 
efficiently configure the HPC computations. MLAutoTuning can be applied at
multiple distinct points, and can be used for a range of tuning and
optimization objectives. For example: (i)  mix of performance and quality of
results using parameters provided by learning
network~\cite{MicrosoftSummit2018A,Jeff_Dean2017,Kraska2018,NanoIJHPCA,
botu2015adaptive}; (ii) choose the best set of ``computation defining
parameters" to achieve some goal such as providing the most efficient training
set with defining parameters spread well over the relevant phase
space\cite{Gavrishchaka2019-bm,Matkovic2018-yy}; (iii) tuning model parameters
to optimize model outputs to available empirical
data~\cite{Gentzsch2018-eo,Ozik2019-ce,JCS_Kadupitiya_Geoffrey_C_Fox_and_Vikram_Jadhao2018-qf,Spellings2018-it}.

 \subsection{MLafterHPC} ML analyzing results of HPC as in trajectory analysis
and structure identification in biomolecular simulations
\cite{Beckstein2018BDEC}.

\section{MLaroundHPC Classification and Exemplars} 

The interaction between models and simulation data occurs in two directions:
(i) The problem of how to use multi-modal data to inform complex models in the
presence of uncertainty, and (ii) How, where, when, and from which source to
acquire simulation data to optimally inform models with respect to a
particular goal or goals is fundamentally an optimal experimental design
problem. Creating the conceptual and technological framework in which models
optimally learn from data and data acquisition is optimally guided by models
presents significant challenges systems of interest are complex, multiscale,
strongly interacting/correlated, and uncertain.

It is important to separate the modes and mechanics of how learning is
integrated with HPC simulations, from the functional motivations of doing so.
Based upon an extensive analysis of the current state of the field, the three
primary modes and mechanisms for integrating learning with HPC simulations
are--- substitution, assimilation and control. Each represents a broad range
of subcategories and possibilities, which no doubt will change rapidly as the
state of theory and practice evolves.

Independent of the modes and mechanisms of integration, we find that there are
three functional drivers of the integration:

\begin{asparaenum}

\item {\bf Improving Simulations:} The essence of this driver is to use
learning to configure and select simulations effectively. There are several
approaches to learning the configuration of physical system being studied,
ranging from improving the learning models using simulation data dynamically,
to using models to determine simulation configurations and/or parameters, as
well as possibly learn configurations of system and software for improved
performance on particular hardware and input parameters~\cite{KASSON201887}.

\item {\bf Learn Structure, Theory and Model for Simulation:} Here the
simulations are used to gradually improve the model or theory, which are in
turn used to improve simulations in some fashion (e.g., as per previous
point). As simulations proceed the model learns the structure or even
underlying principles, and is gradually refined either by coarse-graining or
using better approximations to the effective
potentials~\cite{Chiavazzo2017-gj}.

\item {\bf Learn to make Surrogates: } An increasingly common and important
driver is the use of ML (which are often deep networks) to learn the function
representing the output of the simulation. Such learned representations also
often referred to as surrogates, can be used to determine either the
parameters or the effective ``fields''~\cite{Gentzsch2018,Han2018-il}.

\end{asparaenum}

It is no surprise that the first driver is the most widely investigated and
applied; the rate of progress in the second and third drivers is rapid and
impressive.  We now discuss the primary modes and mechanisms in which the
above three scenarios are often implemented.


\subsection{Substitution}

In this mode, a surrogate model is used to substitute an essential element of
the original simulation (method). The surrogate model is used to create
multi-scale or coarse grained surrogate modeling, which could either learn the
structure or theory of original simulation.



\subsubsection*{Example} Roitberg \textit{et al.} \cite{s.smith_ani-1:_2017}
trained a network on using fine grained Quantum Mechanical DFT calculations.
The resulting ANI-1 model was shown to be chemically accurate, transferrable,
with a performance similar to a classical force field, thus enabling ab-initio
molecular dynamics at a fraction of the cost of ``true'' DFT ab-initio
simulations. Extensions of their work with an active learning (AL) approach
demonstrated that proteins in an explicit water environment can be simulated
with a NN potential at DFT accuracy~\cite{smith_less_2018}. 

In general the focus has been on achieving DFT-level accuracy because NN
potentials are not cheaper to evaluate than most classical empirical
potentials. However, replacing solvent-solvent and solvent-solute
interactions, which typically make up 80\%-90\% of the computational effort in
a classical all-atom, explicit solvent simulation, with a NN potential
promises large performance gains at a fraction of the cost of traditional
implicit solvent models and with an accuracy comparable to the explicit
simulations\cite{Wang:2018aa}.

\subsection{Assimilation}

In this mode, data from simulations, offline external constraints, or
real-time experiments are integrated into physics-based models, which are then
assimilated into traditional simulations. The canonical examples are improving
the Hamiltonian or Force Fields, or in classical data assimilation studies
such as in climate and weather prediction, where in data assimilation involves
continuous integration of time dependent simulations with observations to
correct the model, which are combined  and updated with traditional simulation
model.

\subsubsection*{Example} Current climate models are too coarse to resolve
many of the atmosphere’s most important processes. Traditionally, these
subgrid processes are heuristically approximated in so-called
parameterizations. However, imperfections in these parameterizations,
especially for clouds, have impeded progress toward more accurate climate
predictions for decades. Cloud resolving models alleviate many of the gravest
issues of their coarse counterparts but will remain too computationally
demanding for climate change predictions for the foreseeable future. In
Ref.~\cite{Rasp9684}, a deep neural network is trained to represent all
atmospheric subgrid processes in a climate model by learning from a multiscale
model in which convection is treated explicitly. The trained neural network
then replaces the traditional subgrid parameterizations in a global general
circulation model in which it freely interacts with the resolved dynamics and
the surface-flux scheme. The prognostic multiyear simulations are stable and
closely reproduce not only the mean climate of the cloud-resolving simulation
but also key aspects of variability, including precipitation extremes and the
equatorial wave spectrum. Furthermore, the neural network approximately
conserves energy despite not being explicitly instructed to.
Ref.~\cite{Rasp9684} uses deep learning to leverage the power of short-term
cloud-resolving simulations for climate modeling; the approach is fast and
accurate, thereby showing the potential of machine-learning–based approaches
to climate model development.

\subsection{Control and Adaptive Execution} In this mode, the simulation
(or ensemble of simulations) are controlled  towards important and interesting
parts of simulation phase space. Sometimes this involves determining the
parameters of the next stage (iteration) of simulations based upon
intermediate data. Sometimes the entire campaign can be adaptively steered
towards an objective, which in turn could involve getting better data via
active learning based upon an objective function, or use a policy-based
reinforcement learning approach to steer the computational campaign.

\subsubsection*{Example} A fundamental problem that currently pervades
diverse areas of science and engineering is the need to design expensive
computational campaigns (experiments) that are robust in the presence of
substantial uncertainty. A particular interest lies in effectively achieving
specific objectives for systems that cannot be completely identified. For
example, there may be “big data” but the data size may still pale in
comparison with the complexity of the system, or the available data may be
scarce due to the prohibitive cost of experiments.

A framework for the objective driven experiment design (ODED) will support the
integration of scientific prior knowledge on the system with data generated
via simulations, quantify the uncertainty relative to the objective, and
design optimal experiments that can reduce the uncertainty and thereby
directly contribute to the attainment of the objective.

\section{MLAroundHPC CyberInfrastructure}

We distill the analysis and description of MLAroundHPC modes and examples into
three cyberinfrastructure categories: (i) algorthms, benchmarks and methods;
(ii) system software and runtime, and (iii) hardware.

\subsection{Algorithms, Benchmarks and Methods}

The methodologies by which experiments inform theory, and theory guides
experiments, remain ad hoc, particularly when the physical systems under study
are multiscale, large-scale, and complex. Off-the-shelf machine learning
methods are not the answer; these methods have been successful in problems for
which massive amounts of data are available and for which a predictive
capability does not rely upon the constraints of physical laws. The need to
address this fundamental problem has become urgent, as computational campaigns
at pre-exascale, and soon exascale, will entail models that span wider ranges
of scales and represent richer interacting physics. Open issues and
research questions include:

\begin{asparaenum}

\item Does the crossover point --- at which prediction based approaches are
better than traditional HPC simulations, suggest or motivate a need to
redesign some simulation algorithms so that MLforHPC effective? Similarly, if
HPC simulations are going to serve as important sources of data generation, is
there an opportunity to devise novel learning algorithms and methods so as to
support more effective MLforHPC?

\item Simulations are simply 4D time-series data! Thus, there ought to be important analogies between time series ML research and MLforHPC.

\item Importance of canonical problems: Understanding of which learning
methods work, why and for which problems. How do we develop  benchmarks to
highlight different application and system features? By extension, how do we
develop proxy apps to represent the applications?

\item Understanding Performance: What are the performance metrics that
represent the integrated working of learning and simulations? What is the
comparison in scientific discovery between the large increase in performance
possible with true exascale machines and the exascale (or zettascale)
“effective performance” possible with MLforHPC? How does the interplay of raw
performance and effective performance influence the mapping of applications to
compute systems?

\end{asparaenum}



\subsection{System Software and ML-HPC Runtime Systems:}

MLforHPC needs to support large scale simulations and learning, and their
integrated and concurrent execution. The combined workload --- distinct ML and
HPC computation tasks, will need to be run flexibly. For example, sometime the
HPC simulation will be used to generate training data and then run ML;
sometimes the ML will be responsible for inference as HPC simulations are
generating data. On occasions, HPC simulations will run after Learning (or
vice versa), but sometimes they will be intertwined in a single job. Thus, it
is imperative to understand the general control and coupling between Learning
elements (L), HPC Simulation (S). In many cases a third general component ---
experiments or observations (E) may also be needed.

There are several dimensions to characterize the coupling between these
components, including temporal and data volumes. The former will determine the
type of algorithms and learning approaches taken; the latter software scaling
and performance requirements. Furthermore, the specific type of coupling could
yield steering or control. (Component X is said to steer component Y, when X
provides the relevant information to determine the execution of Y. Whether Y
accepts or not, is determined by additional considerations such as objective,
policy, etc. Steering is a necessary condition for control; not all steering
represents control).

Different scenarios for coupling – information and control flow, between
different elements E, L and S. Scenario III covers two possibilities: learning
element controls experimental data source, or Simulation controls experimental
data source. The logical coupling disregards the physical location of the
elements, e.g., E could be on an Edge device or a HPC cluster.

In order to support the real-time application requirements, it is important to
achieve near real-time training and prediction to control or steer S or E. For
example, build low dimensional representation of states from trajectory
analysis. The strong scaling of just L is inadequate, and scaling properties
of integrated L + S elements are needed for MLaroundHPC applications. A
preliminary analysis suggests that this can be achieved by adapting the ratio
of the cardinality of L, S and E, viz., N$_L$ (the number of learning) to
N$_S$ (number of simulation elements) being time-dependent . These translate
into support for coordinated execution of a large number of concurrent and
heterogeneous simulations as well as enabling adaptive execution and resource
partitioning between simulation and learning elements.

Additional considerations that a runtime system to support the concurrent
execution of ML and HPC elements include: Is a single run-time system possible
that will be able to support the different classes of MLforHPC, varied data
rates (from trivial to O(100)GB/s) and latency tolerance (from $<$ O(1)s to
O(100)s)? Can a single runtime system support the full range of fine-grained
to coarse-grained coupling between learning and simulation components? What
are the considerations and constraints that inform performance guarantees and
workload balancing (e.g., dynamically varying the number of learning elements
and simulation elements)?

\subsection{Hardware and Platform Configuration}

What fraction of time (resource) is spent in ML component and how does this
change with scale? What is the frequency and extent of coupling between
learning and simulations? Insight into the above questions could influence
optimal architecture, e.g., when the ratio of learning (training and
inference) is small, a classic supercomputing architecture linked to a
separate learning system might be acceptable if not optimal. Conversely, when
the ratio is large, a tightly integrated system supercomputer might be more
suitable? What are the quantitative determinants of an optimal platform? How
should a balanced system across a range of MLaroundHPC applications be
designed: fixed dollars for learning vs simulations, or a dollar distribution
that tracks the relative computing intensity? Or one that optimizes inference
phase versus training phase? Should future HPC platforms be designed to
support both phases, or is platform specialization for training and inference
most effective?

Hardware and platform considerations that arise from uncertainty in technology
roadmap and pricing include:

\begin{asparaenum}

\item Role and importance of heterogeneous accelerators,
especially as a new generation of ML accelerators is developed  that may not
be in simulations (currently GPU accelerators often useful in both ML and
simulation); (ii) As we expect time series in data assimilation likely to use
RNN’s and the importance and pervasiveness of RNN to increase, when should
Recurrent neural networks RNN (commonly used in learning sequences) need
different accelerators from convolutional neural nets)?

\item Requirements also suggest the need for fast I/O and internode
communication to enable ML and Computation to run together and exchange
information with each other and with sources of streaming data. It is not
evident how large and fast disks should be organized, but disks on each node
seem required to hold data to be exchanged between simulation and ML
components of a job and for accumulating training data and NN weights.

\item Need ML optimization and Simulation optimization spread through 
machine and fast ways for ML and simulation to exchange data. Given the
emergence of cloudlets (aka fog computing), there is a need to support
HPC/Cloud, Fog and Edge platforms, as well as their integration.

\end{asparaenum}


\section{Discussion and Conclusion}

\noindent 
The state of HPC in 2020 presents challenges and opportunities. On the one
hand, HPC methods and platforms are becoming pervasive and necessary for
scientific advances. On the other, traditional HPC computations are reaching
various limits. The implications of hardware and architectural trends are well
known: the end of Dennard scaling and of Moore's Law as originally formulated,
is yielding very different processor architectures; achieving performance
gains is becoming harder, while requiring significant, if not unsustainable
software investment and algorithmic reformulation.

The HPC community has --- somewhat naively, assumed that as long as
performance gains from hardware are possible, traditional simulation based
methods will continue to provide increased scientific insight. However,
without careful examination of the scientific efficiency or effective
performance of existing simulation and first principles methods, it is not
obvious that traditional simulations represent the optimal approach at
exascale and beyond, and on subsequent generation of supercomputers. In other
words, we may be reaching limits of both hardware and methodological
performance gains.

There is a need for major functionality and performance increases that are
\textbf{independent of changes in hardware}. In traditional HPC the prevailing
orthodoxy ``Faster is Better'' or what is worse, the conflation of ``bigger''
with ``better'' has driven the quest for hierarchical parallelism to speeding
up single units of works. Relinquishing the orthodoxy of hierarchical
parallelism as the primary route to performance is necessary. In fact, there
is a need to carefully reconsider discredited approaches, while adopting the
new paradigms.

Enter ``Learning Everywhere"  --- the essential idea of which, is that by
embedding learning methods and approaches in all aspects of the system
configuration and application execution, the effective performance can be
dramatically improved.

There is a regime where learning based predictive approaches are going to
outperform first-principles and simulation-based approaches. The exact sweet
spot or crossover point is non-trivial to determine: it will be application
specific, depend upon complexity of learned models, volume and cost of data,
as well as effectiveness and cost of simulations, inter alia. However, the
underlying idea that surrogate learned models will represent effective
performance improvements over traditional approaches, is a powerful one, and
is an important generalization of the multi-scale, coarse-grained approaches
used in many physical sciences.

Learning Everywhere is one specific example of the paradigmatic shift in
scientific computing that will be needed at extremes scales. Statistical
computing, which incorporates elements of approximate computing, uncertainty
minimization and other objective driven dynamic computational campaigns will
substitute predefined ``static'' computational campaigns. Nowhere is the
impact of this likely to be greater than in those domains which require the
assimilation of streaming and dynamic data, or computational campaigns that
are statistical in nature and driven towards optimality or objectives. These
methodological innovations will heighten the importance of adaptive execution
of ensembles of heterogeneous models, and will require novel scalable
middleware systems.

{\bf Looking Ahead: } The pace of innovation in learning for science is
intense and rapid. No surprises it is difficult to predict the exact
trajectory or state for anything but the immediate future. It is safe to
expect major impact of ML on science in essentially all areas and in multiple
modes: many traditional physics applications including simulations and Monte
Carlo methods are being reformulated using learning
approaches~\cite{Mehta2019-rm}.

Integrating learning with HPC provides an opportunity to enhance methods and
for some domains such as molecular science to jump ahead. For other field,
such as high-energy physics, that did not invest and anticipate the
disruptions arising from end of Dennard and Moore's law resulting in the
explosion of heterogeneous computing and accelerators, it presents an
important opportunity to simply by-pass and leapfrog a generation of
simulation enhancements!

Impressive, if not inspiring papers that apply learning to societally
important problems such as climate change~\cite{Rolnick2019-bm} are valuable
harbingers. Molecular sciences has been an enthusiastic adopter of learning
methods: Machine Learning used in materials simulation to aid the design of
new materials and to understand properties~\cite{Tabor2018-mh}; predict
reaction coordinates~\cite{Brandt2018-xt}; and enhanced
sampling~\cite{Chen2017-mr} and dynamics on long
time-scales~\cite{Noe2018-qh}. Even as the use of ML in science changes,
important advances in the way ML is formulated are happening. For example,
Ref.~\cite{Tan2019-xd} shows how to scale CNN’s as problems scale --- which
will be crucial in using NN for complex physics systems. In fact,
Ref.~\cite{Chen2018-hp} uses ODE’s to build a continuous neural network rather
than one built from a set of layers.

Enhancements to ML will be necessary to deliver on new and promising uses of
learning in science, such as the application of DL for time series ---
geospatial and simulation trajecctory data (which are simply 4D time series).
These problems can be formulated as graphs spatially (with convolutional NN’s)
and as sequences (with recurrent NN’s) in time. Many HPC Cloud-Edge systems
will provide such time series, and also reinforce the need for real-time
response which raises difficult trade-offs between performance and
functionality and highlights the role of
HPC~\cite{Chathura_Widanage_Jiayu_Li_Sahil_Tyagi_Ravi_Teja_Bo_Peng_Supun_Kamburugamuve_Dan_Baum_Dayle_Smith_Judy_Qiu_Jon_Koskey_undated-tw}.
In general, the pace of methodological innovation and application requirements
will have important implications for the cyberinfrastructure developed and
deployed for the science of tomorrow.

{\footnotesize{\bf Acknowledgement} This work was partially supported by NSF
CIF21 DIBBS 1443054 and nanoBIO 1720625. SJ was partially supported by
ExaLearn -- a DOE Exascale Computing project.}


\bibliography{draft,nssac,annot,radical,bms,ReferencesTaxonomyPaper}
\bibliographystyle{unsrt}

\end{document}